\documentclass[submission,copyright,creativecommons]{eptcs}
\providecommand{\event}{ICLP 2025 Technical Communication} 

\usepackage{iftex}
\usepackage{graphicx}
\usepackage{booktabs}
\ifpdf
  \usepackage{underscore}         
  \usepackage[T1]{fontenc}        
\else
  \usepackage{breakurl}           
\fi

\usepackage{listings}
\usepackage{xcolor}
\usepackage{amsmath}

\lstdefinestyle{mypython}{
  language=Python,
  basicstyle=\ttfamily\small,
  keywordstyle=\color{blue}\bfseries,
  commentstyle=\color{gray}\itshape,
  stringstyle=\color{orange},
  numbers=left,
  numberstyle=\tiny\color{gray},
  stepnumber=1,
  showstringspaces=false,
  tabsize=4,
  breaklines=true,
  frame=single,
  rulecolor=\color{black!30},
  backgroundcolor=\color{black!5},
  literate=
  {←}{{$\leftarrow$}}1
  {→}{{$\rightarrow$}}1
}

\title{Machine Learning Model Integration with Open World Temporal Logic for Process Automation}
\author{Dyuman Aditya
\institute{Syracuse University, Syracuse, NY USA}
\email{daditya@syr.edu}
\and
Colton Payne
\institute{Arizona State University, Tempe, AZ USA}
\email{crpayne2@asu.edu}
\and
Mario Leiva
\institute{Dept. of Computer Science \& Eng.,\\ Inst. for Computer Science \& Eng.\\ Universidad Nacional del Sur, BA, Argentina}
\email{mario.leiva@cs.uns.edu.ar}
\and
Paulo Shakarian
\institute{Syracuse University, Syracuse, NY USA}
\email{pashakar@syr.edu}
}
\def\titlerunning{Machine Learning Model Integration with Open World Temporal Logic for Process Automation}
\def\authorrunning{D. Aditya, C. Payne, M. Leiva, P. Shakarian}
\begin{document}
\maketitle

\begin{abstract}
Recent advances in Machine Learning (ML) have produced models that extract structured information from complex data. However, a significant challenge lies in translating these perceptual or extractive outputs into actionable and explainable decisions within complex operational workflows. To address these challenges, this paper introduces a novel approach that integrates the outputs of various machine learning models directly with the PyReason framework, an open-world temporal logic programming reasoning engine.
PyReason's foundation in generalized annotated logic allows for the incorporation of real-valued outputs (e.g., probabilities, confidence scores) from a diverse set of ML models, treating them as truth intervals within its logical framework. Crucially, PyReason provides mechanisms, implemented in Python, to continuously poll ML model outputs, convert them into logical facts, and dynamically recompute the minimal model to enable decision-making in real-time. Furthermore, its native support for temporal reasoning, knowledge graph integration, and fully explainable interface traces enables an analysis of time-sensitive process data and existing organizational knowledge. By combining the strengths of perception and extraction from ML models with the logical deduction and transparency of PyReason, we aim to create a powerful system for automating complex processes. This integration is well suited for use cases in numerous domains, including manufacturing, healthcare, and business operations.
\end{abstract}

\section{Introduction}
\label{sec:intro}
Recent advances in Machine Learning (ML) have produced models that extract structured information from complex data. Object detectors, image classifiers, and large language models can now recognize patterns, classify inputs, and extract entities. Their outputs, such as object locations, image labels, or text sentiments, can help automate and improve real-world processes.

However, a significant challenge lies in translating these perceptual or extractive outputs into actionable and explainable decisions within complex operational workflows. ML models, while proficient in their specific tasks, often lack the inherent mechanisms for explainable logical reasoning, transparently managing uncertainty, integrating explicit domain knowledge, or handling the temporal dynamics crucial for many automation scenarios. Bridging this gap requires a framework that can synthesize these varied ML outputs and subject them to rigorous, explainable logical analysis.

\begin{figure}[t]
    \centering
    \includegraphics[width=0.8\linewidth]{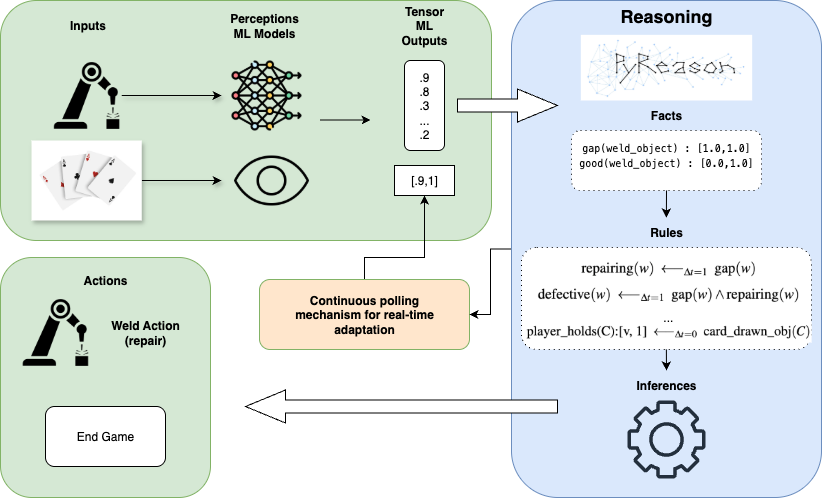}
    \caption{A conceptual overview of the proposed integration framework. (1) Machine learning models provide perceptual outputs (e.g., tensor, probabilities). (2) The PyReason framework converts these outputs into logical facts with truth-value intervals. (3) These facts are used within a temporal logic program to derive actionable, reasoned decisions. The feedback loop illustrates the continuous polling mechanism for real-time adaptation}
    \label{fig:intro}
\end{figure}

To address these challenges, this paper introduces a novel approach that integrates the outputs of various machine learning models directly with the PyReason framework~\cite{adityapyreason2023}, an open-world temporal logic reasoning engine. Figure~\ref{fig:intro} provides a high-level overview of this integration framework. PyReason's foundation in generalized annotated logic~\cite{ks92} directly incorporates real-valued outputs (e.g., probabilities, confidence scores) from diverse ML models, treating them as truth intervals within its logical framework.  We note that the existing work focuses on integration between machine learning models and logic-based reasoning by either modifying the ML model (e.g., see \cite{Badreddine2022,deeprpoblog18,neurasp,cornelio2023learning,GIUNCHIGLIA2024109124,NEURIPS2022c182ec59}) or by adding rules about the failure of the model to the logic program (e.g., see \cite{kricheli24,xi2025rulebasederrordetectioncorrection,leiva2025consistencybasedabductivereasoningperceptual}).  In this paper, we examine practical methods complementary to both approaches - focusing on software engineering questions concerning the integration between logic programs and ML models. This distinction carries a significant impact: by not requiring modifications to the ML model, our approach allows for the seamless integration of pre-trained, ``black-box'' systems. This enhances modularity and practical applicability, as it enables the use of state-of-the-art proprietary models or those that are computationally expensive to fine-tune, thus lowering the barrier for creating complex neuro-symbolic applications.

A central pillar of our proposal is the ability of PyReason to implement logic programs with Python code. This is achieved through mechanisms that enable Python code to continuously poll ML model results, converting these perceptions into logical facts that are directly injected into the logic program. This process not only feeds the reasoning engine, but also triggers the re-computation of the minimal model, ensuring that logical decisions always reflect the latest observations from the ML model. Given the ICLP audience's familiarity with both Python and logic programming, we have prioritized a design that explicitly intersperses both paradigms, facilitating an understanding of the interaction between ML perception and logical deduction.
We have made several design decisions to simplify this integration and improve usability. For example, the introduction of specific base classes such as \lstinline[style=mypython]!LogicIntegrationBase! and \lstinline[style=mypython]!TemporalLogicIntegratedClassifier! standardizes the inference flow and the conversion of predictions into logical facts. These decisions simplify syntax, manage optional arguments, and support a robust, modular approach to complex process automation. 

In addition to this integration, PyReason supports temporal reasoning, knowledge graph integration, and explainable inference traces. While traditional logic programming languages like Prolog can provide explainable reasoning paths, they are less suited for integrating existing knowledge graphs and handling time-based logic. Python, by contrast, offers better support for these needs through libraries like GraphML and NetworkX, and is more adaptable for implementing temporal reasoning. By combining outputs from machine learning models with PyReason's logical reasoning capabilities, we aim to build a system for automating complex processes. This approach is applicable across domains such as manufacturing (using object detection for quality control), healthcare (image classification for diagnostics), and business operations (LLMs for document processing and customer support).

The remainder of this paper is structured as follows. Section\ref{sec:pyreason-intro} introduces the PyReason framework, its logical foundations, and its reasoning capabilities. Section\ref{sec:machine-learning-integration} details the integration with machine learning models via a standardized interface. Section\ref{sec:case-studies} presents two case studies—robotic welding and a card game—to illustrate practical applications. Section\ref{sec:conclusions} concludes with a discussion of the broader implications.

\section{The PyReason Framework}
\label{sec:pyreason-intro}
PyReason is a software framework developed for open-world temporal logic, implemented in Python~\cite{adityapyreason2023}. It is founded on generalized annotated logic, a system which not only encapsulates the current range of differentiable (i.e., fuzzy) first-order logics but also integrates temporal extensions to support inference over finite durations. A primary design objective of PyReason is to provide direct support for reasoning on graphical structures, such as knowledge graphs, social networks, and biological networks. The framework is engineered to generate fully explainable traces of its inference processes, ensuring transparency. It also includes type-checking and a memory-efficient design for scalable deductive inference. PyReason endeavors to supply a modular and extensible platform for deduction, addressing the constraints of bespoke software frequently developed for particular logics within neuro-symbolic frameworks. It captures a breadth of capabilities observed in diverse neuro-symbolic frameworks, encompassing fuzzy logic, open-world reasoning, temporal reasoning, and graph-based reasoning. The PyReason software is publicly accessible via GitHub (\url{https://github.com/lab-v2/pyreason}).

\subsection{Core Logical Principles and Reasoning over Graphs}

The logical underpinning of PyReason is generalized annotated logic, augmented with various extensions to support a wide array of capabilities. A defining characteristic is its application of real-valued interval annotations, enabling logical statements to be annotated with elements from a lower semilattice structure composed of intervals that are subsets of $[0, 1]$. This design aligns with truth intervals in fuzzy operators and paradigms in neuro-symbolic reasoning. Classical logic supports annotations $[0, 0]$ (false) and $[1, 1]$ (true), while tri-valued logic can use $[0, 1]$ for "unknown". Negation of an atom annotated with $[l, u]$ results in an annotation of $[1-u, 1-l]$ for its strong negation.

PyReason integrates temporal logic over finite periods through "interpretations", which are an initial set of facts correct at $t=0$. Temporal annotations $I(A, \hat{t}):[L,U]$ represent discrete time-points, where $\hat{t}$ is a specific time $t$ or $s$ for static interpretations (unchanging over time). Non-static predicates can change annotations over time without a monotonicity requirement. The framework operates under an open-world assumption: interpretations not explicitly defined are initially "unknown" (e.g., $[0,1]$).

The framework is designed to reason about graphical structures, such as knowledge graphs. It assumes a graph $G=(\mathcal{C},E)$, where nodes $\mathcal{C}$ are constants, and edges $E$ specify potential relationships. Predicates are typically unary (node labels) or binary (edge labels). A reserved binary predicate `rel` exists, where $rel(a,b)$ is a tautology if $(a,b) \in E$ and uncertain otherwise. Graphical syntactic extensions allow for operators like requiring the existence of $k$ items.

\subsection{Key Reasoning Capabilities and Features}

Logical rules are the primary mechanism for modifying annotations of non-static predicates. PyReason supports ground rules, universally quantified non-ground rules (for reasoning within or across edges), and rules with rule-based quantifiers in the head. Rules incorporate a temporal component through a time delay, denoted as $\Delta t$. Syntactically, this delay is represented as a subscript to the implication arrow (e.g., $\leftarrow_{\Delta t}$, specifying the number of timesteps that must elapse after the body is satisfied before the head's annotation is updated. This feature explicitly defines the temporal nature of the logic programs used within PyReason. Rule heads often involve functions $f(x_1, ..., x_n)$ (e.g., T-norms, T-conorms, algebraic functions) to combine annotations from body literals.

Deduction centers around a fixpoint operator, $\Gamma$, proven to produce all entailed atoms by a logic program. This computation is exact, yielding the minimal model and allowing for entailment checking. A significant advantage is the fully explainable result, with PyReason providing the series of inference steps for any entailment query.

To mitigate computational complexity, especially the grounding problem, PyReason incorporates predicate-constant type checking. This leverages knowledge graph sparsity, where nodes have types and predicates are defined over specific types. During initialization, ground atoms are created only for compatible predicate-constant pairs, significantly reducing the search space. This is an optional feature.

PyReason detects and manages logical inconsistencies. Inconsistency can occur if a new interpretation $[L',U']$ is not a subset of the current $[L,U]$, or between an atom and its negation, or complementary predicates. PyReason flags and can report these inconsistencies. The explainable fixpoint operator allows tracing the cause. An option exists to resolve inconsistency by resetting the annotation to $[0, 1]$ (uncertainty) and setting the atom to static.

\subsection{Implementation and Neuro-Symbolic Relevance}

PyReason is a modern Python framework designed for scalability and correctness. Graphical input is via PrologML or a Networkx Prolog, with initial conditions and rules in text format. The Numba JIT compiler optimizes key operations for performance and supports CPU parallelism. For memory efficiency, only current interpretations are stored. However, past states are reconstructible via rule traces, which  enable explainability for long-running implementations. This design supports canonical and non-canonical models with minimal computational overhead by recomputing interpretations at each step.

PyReason is strategically positioned as an instrument in neuro-symbolic AI. Generalized annotated logic, its foundation, captures a wide spectrum of real-valued, temporal, and fuzzy logics used in frameworks like Logical Tensor Networks (LTN)  and Logical Neural Networks (LNN), both viewable as subsets of annotated logic. PyReason can conduct inference on logic programs learned or modified by such systems. It enables capabilities like graph-based and temporal reasoning, potentially absent in other neuro-symbolic frameworks' native logics. The explainability of its inference is essential to understand and debug complex neuro-symbolic models. By supporting generalized annotated logic and its extensions, PyReason facilitates system design independent of the learning process, offering a robust deductive engine.

\section{Integration with Machine Learning Models}
\label{sec:machine-learning-integration}


PyReason allows users to integrate machine learning (ML) models directly within the logical framework. ML models output \emph{tensors}, which are converted into PyReason \emph{facts} that can be used to perform inference. This section outlines this process and illustrates its simplicity within our framework. 

\subsection{Logic Integrated Base Class}
PyReason defines an abstract base class \lstinline[style=mypython]!LogicIntegrationBase!, designed for subclassing, to integrate specific machine learning models. This class declares three protected methods that are called sequentially and must be implemented by every subclass:

\begin{enumerate}
    \item \lstinline[style=mypython]!_infer!: Execute the model and return its raw output tensor.
    \item \lstinline[style=mypython]!_postprocess!: Transform the tensor (e.g.\ apply a sigmoid) into interpretable scores.
    \item \lstinline[style=mypython]!_pred_to_facts!: Convert the post-processed predictions into logical facts for downstream reasoning.
\end{enumerate}

\subsection{Logic Integrated Classifier}
This section covers how the \lstinline[style=mypython]!LogicIntegrationBase! class can be used to integrate a simple binary classifier which is then used in the example in Subsection~\ref{subsec:welding-example}. We outline the implementation of the three methods discussed earlier.

\noindent
\textbf{Model inference:}
The \lstinline[style=mypython]!_infer! method is implemented as follows, where the underlying model is called to get the raw logits.
\begin{lstlisting}[style=mypython, caption={Model Inference}]
    def _infer(self, x: torch.Tensor) -> torch.Tensor:
        return self.model(x)
\end{lstlisting}

\noindent
\textbf{Post Processing}
The \lstinline[style=mypython]!_postprocess! method applies a sigmoid function over the raw model outputs to get a probability distribution over the classes.
\begin{lstlisting}[style=mypython, caption={Post-Processing}]
    def _postprocess(self, raw_output: torch.Tensor) -> torch.Tensor:
        logits = raw_output  # Output from the _infer function
        return F.softmax(logits, dim=1)
\end{lstlisting}

\noindent
\textbf{Predictions to Facts}
\label{sec:pred-to-facts}
The \lstinline[style=mypython]!_pred_to_facts! method is responsible for converting the post-processed predictions into logical facts. To do this, we define a set of user interface options:

\begin{enumerate}
    \item \lstinline[style=mypython]!threshold!: Beyond which predictions should be integrated
    \item \lstinline[style=mypython]!set_lower_bound!: Whether to modify the lower bound of the resulting fact
    \item \lstinline[style=mypython]!set_upper_bound!: Whether to modify the upper bound of the resulting fact
    \item \lstinline[style=mypython]!snap_value!: The value to set the upper or lower bound (depending on the setting) of the resulting fact. If this is left empty, the original output from the post-processing function will be used.
\end{enumerate}

\begin{figure}[t]
    \centering
    \includegraphics[width=0.7\linewidth]{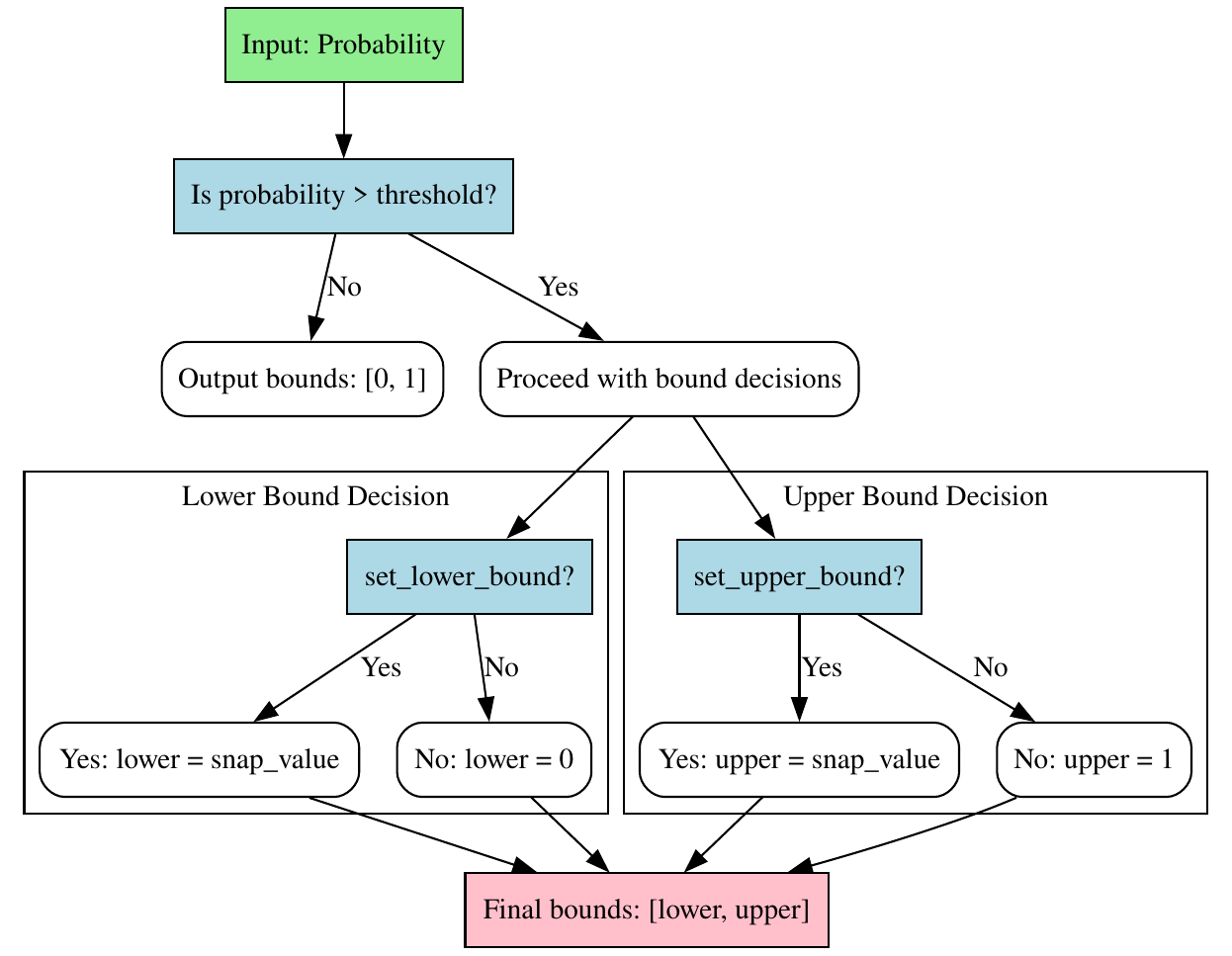}
    \caption{Conversion from probabilities to annotated bounds}
    \label{fig:interface-options}
\end{figure}

For simplicity, we outline the process of converting probabilities to annotated bounds as a flowchart depicted in Figure~\ref{fig:interface-options}. By following this logic, we are able to convert model output tensors into annotated facts which can be added into PyReason for downstream inference.

\subsection{Automated Temporal Logic Integrated Classifier}\label{sec:automatic-temporal-classifier}
In many real‐world applications, data arrives continuously from sensors or cameras. Logical inferences must adapt in real time. To address this, PyReason provides the \texttt{TemporalLogicIntegratedClassifier}, a subclass of \texttt{LogicIntegratedClassifier} that ties PyReason’s internal temporal reasoning to wall‐clock time. This class adds two key parameters: \texttt{poll\_interval}, which specifies the frequency (in seconds or PyReason time) to poll the model with new input data if the \texttt{poll_condition} is either satisfied or not defined; and \texttt{poll_condition}, a PyReason query that determines whether a new inference cycle should be triggered.

\begin{figure}[t]
    \centering
    \includegraphics[width=0.6\linewidth]{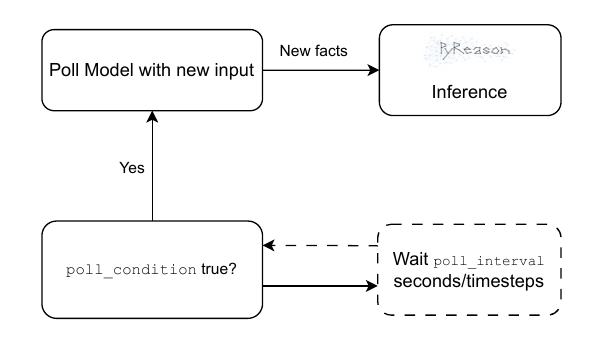}
    \caption{Background model-polling and inference loop}
    \label{fig:temporal-classifier-integration}
\end{figure}

At each interval, PyReason evaluates the \texttt{poll\_condition}. If the condition is satisfied, or if no condition is specified, the system automatically:
\begin{enumerate}
  \item Loads new input (e.g.\ an image or sensor reading).
  \item Invokes the underlying ML model to produce raw outputs.
  \item Converts those outputs into annotated facts (as shown in Section~\ref{sec:machine-learning-integration}).
  \item Runs the fixpoint operator to update the interpretation given the new facts.
\end{enumerate}
This end‐to‐end loop (Figure~\ref{fig:temporal-classifier-integration}) requires no additional glue code: PyReason handles the timing, model polling, fact injection, and inference in a single, explainable pipeline.

Section~\ref{sec:case-studies} illustrates two concrete workflows built on this machinery: defect detection in robotic welding and decision‐making in a simulated card game. In each case, the temporal classifier ensures that as soon as new data are available, the logic program reflects the latest observations and derives the appropriate conclusions.

\section{Illustrative Case Studies}\label{sec:case-studies}
To demonstrate the practical application and versatility of integrating machine learning models within the PyReason framework, this section presents two distinct case studies. The first, a robotic welding scenario, illustrates how the system can be applied to industrial automation for real-time quality control. It showcases the use of a temporal logic-integrated classifier to detect weld defects, trigger an automated repair process, and make a final determination of product quality.
The second case study, a simple card game, highlights PyReason's ability to manage uncertainty and guide decision-making over time by periodically integrating outputs from an image classifier and applying annotated logic to the evolving state of the game.

\subsection{Welding Classifier}\label{subsec:welding-example}

In an industrial robotic welding cell, detecting weld defects in real time is essential to maintaining product quality and avoiding costly rework. Here, we apply a pre-trained weld-defect detector to identify gaps in each weld image, feed those results into PyReason to attempt automated gap closure, and mark the part as defective if the gap cannot be closed.

\subsubsection{Logical Encoding in PyReason}
We use the automated temporal logic integrated classifier described in Section~\ref{sec:automatic-temporal-classifier} to integrate a welding defect detection model with a logic program that consists of two simple rules:

\[
\text{repairing}(w) \;\longleftarrow_{\Delta t=1}\; \text{gap}(w)
\]
\[
\text{defective}(w) \;\longleftarrow_{\Delta t=1}\; \text{gap}(w)\wedge \text{repairing}(w)
\]

\begin{figure}[t]
    \centering
    \includegraphics[width=0.7\linewidth]{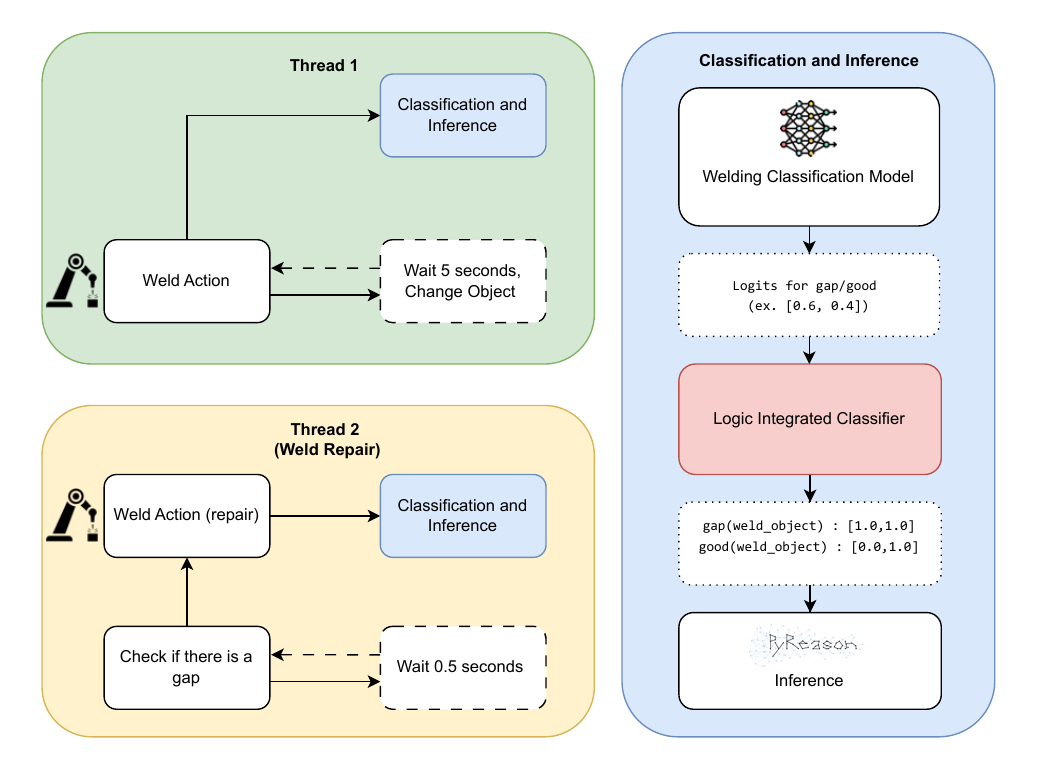}
    \caption{Welding example flowchart}
    \label{fig:welding-flowchart}
\end{figure}

The $\texttt{gap(\text{weld_object})}$ atom is added to the program by PyReason's \texttt{LogicIntegratedClassifier}. These two rules help determine if the weld was unsuccessful and if after a repair was attempted there is still a gap, then the part is marked as defective. A flowchart of the pipeline can be seen in Figure~\ref{fig:welding-flowchart}. The repair and conclusion that a part might be defective run in a background routine (Thread 2) while the main thread (Thread 1) goes over each object to be welded and performs the welding task.

We use the \texttt{TemporalLogicIntegratedClassifier} defined in Section~\ref{sec:automatic-temporal-classifier}, with a \texttt{poll_interval} set to 0.5 seconds and the \texttt{poll_condition} set to \texttt{gap(weld_object):[1,1]}. This allows us to check whether there is a gap in the weld every 0.5 seconds and attempt to repair the gap in the background. 

\subsubsection{Interpreting the Inference Trace}
Table~\ref{tab:welding-node-traces} showcases an explainable PyReason trace after welding 5 objects over 7 fixed-point operations (FPO). It can be seen that both thread 1 and thread 2 update the interpretation simultaneously. Thread 1 is always called first on every new object, and thread 2 is invoked only when a repair is necessary. In FPOs 2-3, a repair is attempted due to a gap being detected in FPO 1 and the object is deemed defective after an unsuccessful repair. Similarly, another repair is invoked in FPO 5, but the repair is successful and the part is not marked as defective.

\begin{table}[t]
  \centering
  \small
  \caption{PyReason explainable trace for the welding scenario}
  \label{tab:welding-node-traces}
  \begin{tabular}{clllll}
    \toprule
    \textbf{FPO} & \textbf{Node}          & \textbf{Label}     & \textbf{Old Bound} & \textbf{New Bound} & \textbf{Thread} \\
    \midrule
    \texttt{0}   & \texttt{weld\_object}  & \texttt{good}      & [0.0,1.0]          & [1.0,1.0]          & 1 \\
    \midrule
    \texttt{1}   & \texttt{weld\_object}  & \texttt{gap}       & [0.0,1.0]          & [1.0,1.0]          & 1 \\
    \midrule
    \texttt{2}   & \texttt{weld\_object}  & \texttt{repairing} & [0.0,1.0]          & [1.0,1.0]          & 2 \\
    \texttt{2}   & \texttt{weld\_object}  & \texttt{gap}       & [0.0,1.0]          & [1.0,1.0]          & 2 \\
    \texttt{3}   & \texttt{weld\_object}  & \texttt{defective} & [0.0,1.0]          & [1.0,1.0]          & 2 \\
    \midrule
    \texttt{4}   & \texttt{weld\_object}  & \texttt{gap}       & [0.0,1.0]          & [1.0,1.0]          & 1 \\
    \midrule
    \texttt{5}   & \texttt{weld\_object}  & \texttt{repairing} & [0.0,1.0]          & [1.0,1.0]          & 2 \\
    \texttt{5}   & \texttt{weld\_object}  & \texttt{good}      & [0.0,1.0]          & [1.0,1.0]          & 2 \\
    \midrule
    \texttt{6}   & \texttt{weld\_object}  & \texttt{good}      & [0.0,1.0]          & [1.0,1.0]          & 1 \\
    \midrule
    \texttt{7}   & \texttt{weld\_object}  & \texttt{good}      & [0.0,1.0]          & [1.0,1.0]          & 1 \\
    \bottomrule
  \end{tabular}
\end{table}

\subsection{Simple Card Game}\label{subsec:cardgame-example}
This case study demonstrates how PyReason's open-world temporal logic can drive decision-making in a simple "42" card game by periodically integrating the output from an image classifier and applying annotated logic to manage uncertainty over time. The objective of the game is to draw as many cards as possible without exceeding a point total of 42. Card values are defined as 3 points for an Ace, 6 for numeric cards, and 9 for face cards. At each turn, a decision is made to draw another card if there is a nonzero probability of remaining at or below the 42-point limit, based on the cards remaining in the deck. The game concludes when any subsequent draw would guarantee exceeding the limit.  

\subsubsection{Logical Encoding in PyReason}
We encode the game state using five predicates and three rules:

\paragraph{Predicates}
\begin{itemize}
  \item \lstinline[style=mypython]!deck_holds(C, D)!: links each card $C$ with the deck $D$ as edges in a graph.
  \item \lstinline[style=mypython]!card_drawn_obj(C)!: classifier output indicating that card $C$ was drawn.
  \item \lstinline[style=mypython]!player_holds(C)!: assigns a fuzzy interval score to $C$ based on \texttt{card\_drawn\_obj(C)}.
  \item \lstinline[style=mypython]!hand_as_point_vals(V)!: total value of held cards, represented as a fuzzy interval.
  \item \lstinline[style=mypython]!odds_of_losing(P)!: probability that the next draw exceeds 42 points. 
\end{itemize}

\paragraph{Rules}

\[
\text{player\_holds(C):} \text{[v,\ 1]} \;\longleftarrow_{\Delta t=0}\; \text{card\_drawn\_obj}(C)
\]
\[
\text{hand\_as\_point\_vals}(\text{hand}) : \text{append\_hand\_annotation\_fn} \;\longleftarrow_{\Delta t=0}\; \text{player\_holds}(\text{card}):[0.3, 1]
\]
\[
\resizebox{\textwidth}{!}{$
\text{odds\_of\_losing}(\text{hand})\!:\!\text{odds\_of\_losing\_annotation\_fn} \!\longleftarrow_{\Delta t=0}\! \text{hand\_as\_point\_vals}(\text{hand})\!:\![0,1],\! \text{deck\_holds}(\text{card},\text{full\_deck})\!:\![0.3,1]
$}
\]

\begin{itemize}
  \item \lstinline[style=mypython]!player_holds(C):!  When a card $C$ is drawn, its corresponding point value $v$ is used to update the lower bound of \texttt{player\_holds(C)}. Each of the 52 cards has a dedicated rule.
  
  \item \lstinline[style=mypython]!hand_as_point_vals(hand)!: This rule updates the cumulative point value of the hand by appending each drawn card's value. It uses a non-grounded rule with an annotation function (\texttt{append\_hand\_annotation\_fn}) to incrementally build the hand score. The threshold $[0.3,1]$ ensures triggering for any card.

  \item \lstinline[style=mypython]!odds_of_losing(hand):!
      Given the current hand value and remaining cards in the deck, this rule estimates the probability of exceeding 42 points on the next draw. The calculation is performed by the \texttt{odds\_of\_losing\_annotation\_fn}, using the fraction of risky remaining cards.
\end{itemize}

\begin{figure}[t]
  \centering
  \includegraphics[width=0.7\linewidth]{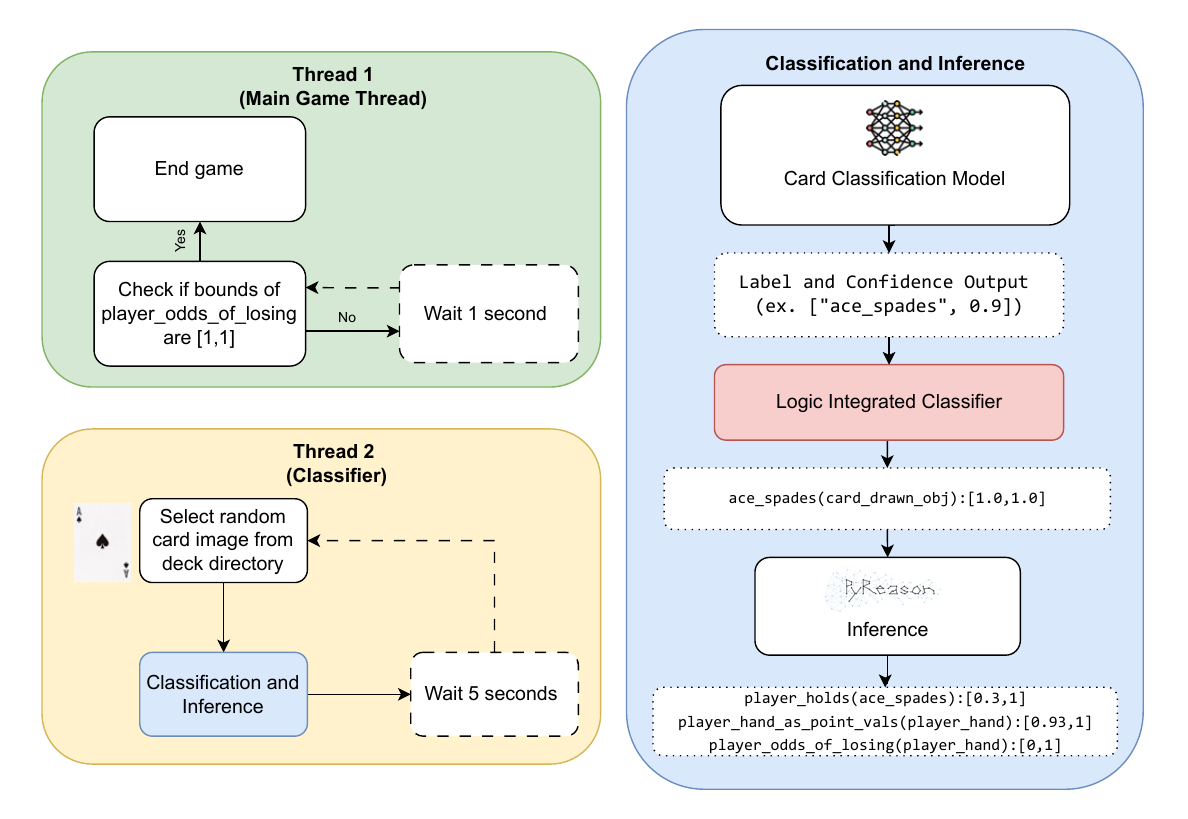}
  \caption{Workflow diagram for the card game simulation.}
  \label{fig:card-game-workflow}
\end{figure}

Figure~\ref{fig:card-game-workflow} depicts the full simulation loop: one thread monitors termination conditions, while another periodically classifies drawn cards and updates the logical state. This concurrent design shows how PyReason’s fixed-point inference can be driven by — and synchronized with — real (wall-clock) time. The classification and inference module periodically polls the image classifier to identify each drawn card and then encodes those results as PyReason facts, allowing for logical reasoning about the evolving game in the background.

\subsubsection{Interpreting the Inference Trace}

The explainable trace in Table 2 records every Fixed‐Point Operation (FPO) in one run of the “42” game, showing how PyReason:

\begin{enumerate}
  \item Asserts each drawn card via the \texttt{card\_drawn\_obj} fact at certainty \([1,1]\).  
  \item Fires \texttt{player\_holds} to add the card’s point value \(v\) to the lower bound of \texttt{hand\_as\_point\_vals.}  
  \item Fires \texttt{odds\_of\_losing} to compute the interval \([p,1]\), where \(p\) is the fraction of remaining cards that would push the total above 42.  When \([p,1]=[1,1]\), no safe draws remain and the game ends.
\end{enumerate}



\begin{table}[h]
  \centering
  \caption{Simplified node traces for one simulation of the card game}
  \label{tab:node-traces-simplified}
  \small
  \begin{tabular}{lllll}
    \toprule
    \textbf{FPO} & \textbf{Node}               & \textbf{Label}                     & \textbf{Old Bound} & \textbf{New Bound} \\
    \midrule
    \texttt{0}   & \texttt{card\_drawn\_obj}  & \texttt{two\_clubs}                        & \texttt{[0.0,1.0]} & \texttt{[1.0,1.0]}   \\
    \texttt{1}   & \texttt{two\_clubs}               & \texttt{player\_holds}             & \texttt{[0.0,1.0]} & \texttt{[0.6,1.0]}   \\
    \texttt{2}   & \texttt{player\_hand}      & \texttt{hand\_as\_point\_vals} & \texttt{[0.0,1.0]} & \texttt{[0.6,1.0]}   \\
    \texttt{3}   & \texttt{player\_hand}      & \texttt{odds\_of\_losing}  & \texttt{[0.0,1.0]} & \texttt{[0.0,1.0]}   \\
    \midrule
    \texttt{4}   & \texttt{card\_drawn\_obj}  & \texttt{ten\_hearts}                       & \texttt{[0.0,1.0]} & \texttt{[1.0,1.0]}   \\
    \texttt{5}   & \texttt{ten\_hearts}            & \texttt{player\_holds}             & \texttt{[0.0,1.0]} & \texttt{[0.6,1.0]}   \\
    \texttt{6}   & \texttt{player\_hand}      & \texttt{hand\_as\_point\_vals} & \texttt{[0.0,1.0]} & \texttt{[0.66,1.0]} \\
    \texttt{7}   & \texttt{player\_hand}      & \texttt{odds\_of\_losing}  & \texttt{[0.0,1.0]} & \texttt{[0.0,1.0]}   \\
    \midrule
    \texttt{8}   & \texttt{card\_drawn\_obj}  & \texttt{six\_clubs}                       & \texttt{[0.0,1.0]} & \texttt{[1.0,1.0]}   \\
    \texttt{9}   & \texttt{six\_clubs}                & \texttt{player\_holds}             & \texttt{[0.0,1.0]} & \texttt{[0.6,1.0]}   \\
    \texttt{10}   & \texttt{player\_hand}      & \texttt{hand\_as\_point\_vals} & \texttt{[0.6,1.0]} & \texttt{[0.666,1.0]} \\
    \texttt{11}   & \texttt{player\_hand}      & \texttt{odds\_of\_losing}  & \texttt{[0.0,1.0]} & \texttt{[0.0,1.0]}   \\
    \midrule
    \texttt{12}   & \texttt{card\_drawn\_obj}  & \texttt{four\_clubs}                        & \texttt{[0.0,1.0]} & \texttt{[1.0,1.0]}   \\
    \texttt{13}   & \texttt{four\_clubs}                & \texttt{player\_holds}             & \texttt{[0.0,1.0]} & \texttt{[0.6,1.0]}   \\
    \texttt{14}  & \texttt{player\_hand}      & \texttt{hand\_as\_point\_vals} & \texttt{[0.666,1.0]} & \texttt{[0.6666,1.0]} \\
    \texttt{15}  & \texttt{player\_hand}      & \texttt{odds\_of\_losing}  & \texttt{[0.0,1.0]} & \texttt{[0.0,1.0]}   \\
    \midrule
    \texttt{16}  & \texttt{card\_drawn\_obj}  & \texttt{jack\_spades}                        & \texttt{[0.0,1.0]} & \texttt{[1.0,1.0]}   \\
    \texttt{17}  & \texttt{jack\_spades}                & \texttt{player\_holds}             & \texttt{[0.0,1.0]} & \texttt{[0.9,1.0]}   \\
    \texttt{18}  & \texttt{player\_hand}      & \texttt{hand\_as\_point\_vals} & \texttt{[0.6666,1.0]} & \texttt{[0.66669,1.0]} \\
    \texttt{19}  & \texttt{player\_hand}      & \texttt{odds\_of\_losing}  & \texttt{[0.0,1.0]} & \texttt{[0.0,1.0]}   \\
    \midrule
    \texttt{20}  & \texttt{card\_drawn\_obj}  & \texttt{ace\_clubs}                        & \texttt{[0.0,1.0]} & \texttt{[1.0,1.0]}   \\
    \texttt{21}  & \texttt{ace\_clubs}                & \texttt{player\_holds}             & \texttt{[0.0,1.0]} & \texttt{[0.3,1.0]}   \\
    \texttt{22}  & \texttt{player\_hand}      & \texttt{hand\_as\_point\_vals} & \texttt{[0.66669,1.0]} & \texttt{[0.666693,1.0]} \\
    \texttt{23}  & \texttt{player\_hand}      & \texttt{odds\_of\_losing}  & \texttt{[0.0,1.0]} & \texttt{[0.23913,1.0]} \\
    \midrule
    \texttt{24}  & \texttt{card\_drawn\_obj}  & \texttt{three\_hearts}                        & \texttt{[0.0,1.0]} & \texttt{[1.0,1.0]}   \\
    \texttt{25}  & \texttt{three\_hearts}               & \texttt{player\_holds}             & \texttt{[0.0,1.0]} & \texttt{[0.6,1.0]}   \\
    \texttt{26}  & \texttt{player\_hand}      & \texttt{hand\_as\_point\_vals} & \texttt{[0.666693,1.0]} & \texttt{[0.6666936,1.0]} \\
    \texttt{27}  & \texttt{player\_hand}      & \texttt{odds\_of\_losing}  & \texttt{[0.23913,1.0]} & \texttt{[1.0,1.0]}   \\
    \bottomrule
  \end{tabular}
\end{table}

\section{Conclusions}
\label{sec:conclusions}
This paper presented a framework to bridge the gap between machine learning perception and logical reasoning by integrating ML models directly with the PyReason temporal logic engine. Our approach uses generalized annotated logic to convert real-valued model outputs into logical facts. It provides a standardized architecture via the \lstinline[style=mypython]!LogicIntegrationBase! and \lstinline[style=mypython]!TemporalLogicIntegratedClassifier! classes, creating a modular, real-time pipeline that requires no additional glue code.

The utility of the framework was demonstrated in two case studies: a welding scenario for real-time process control and a card game for managing uncertainty in decision making. These examples highlighted the system's ability to handle temporal dynamics and produce fully explainable inference traces, which are crucial for debugging and policy refinement.

Beyond the integration with machine learning models for perception, the PyReason framework is architecturally designed to support the integration of external, specialized reasoning engines, a concept analogous to the pluggable constraint solvers found in Constraint Logic Programming (CLP). This modularity allows the logic program to function as a high-level orchestrator that defines a problem space and coordinates with specialized computational modules. While the core of this paper focuses on integrating machine learning models for perception, the framework's underlying design is more general, enabling interfaces with various external solvers. A clear demonstration of this capability is presented in the work~\cite{patil2025reasoningmedicaltriageoptimization}, which describes using PyReason to manage medical triage optimization. In their approach, the logic program dynamically formulates optimization problems by reasoning over situational facts to select the appropriate objectives and constraints. These problem instances are then passed to a dedicated external integer programming solver for resolution. Crucially, the solver's output is translated back into logical facts and reintegrated into the PyReason program, which then performs further symbolic reasoning on these results to guide decision-making. This workflow exemplifies a general pattern where PyReason orchestrates a process by leveraging an external, ``pluggable'' engine for a specialized reasoning task, while maintaining a single, explainable logical framework.

Future work will focus on extending this integration in several key directions. In particular, we look to integrate ``metacognitive'' learned about the failure of the model to improve results on down-stream tasks (e.g., see \cite{kricheli24}).  In particular, early results have shown that the combination of such rules with abductive inference in PyReason can lead to improved ensembling of object detection ML models~\cite{leiva2025consistencybasedabductivereasoningperceptual}.  Based on those results, we look to examine further practical integrations for ML-logic programming integration with such error correction capabilities.  Additionally, we plan to apply the framework to more complex, large-scale industrial scenarios and explore its use with a broader range of ML models, such as Large Language Models for business process automation, to further validate its robustness and versatility.

\section*{Acknowledgments}
This research was supported by the Defense Advanced Research Projects Agency (DARPA) under Cooperative Agreement No. HR00112420370 (MCAI). The views expressed in this paper are those of the authors and do not necessarily reflect the official policy or position of the U.S. Army, the U.S. Department of Defense, or the U.S. Government.

\nocite{*}
\bibliographystyle{eptcs}
\bibliography{bibliography}
\end{document}